\documentclass[conference]{IEEEtran} 

\IEEEoverridecommandlockouts                              

\usepackage{cite}
\usepackage{amsmath,amssymb,amsfonts}
\usepackage{algorithmic}
\usepackage{algorithm}
\usepackage{graphicx}
\usepackage{textcomp}
\usepackage{xcolor}
\usepackage{threeparttable}
\usepackage{multirow}
\usepackage{booktabs}
\usepackage{multirow}
\usepackage{hyperref}

\hypersetup{%
 hidelinks,
 colorlinks=false}

\usepackage{booktabs}   
\usepackage{siunitx}    
\newcommand{\RED}[1]{\textcolor{black}{#1}}

\title{
Strouhal-Aware Model Predictive Control for Efficient Multi-Fin Flapping Locomotion
}

\author{
\IEEEauthorblockN{Yuya Hamamatsu$^{1}$ \quad Zixi Chen$^{2}$ \quad Maarja Kruusmaa$^{1}$ \quad Asko Ristolainen$^{1}$}
\IEEEauthorblockA{$^{1}$Department of Computer Systems, Tallinn University of Technology, Tallinn, Estonia\\
\IEEEauthorblockA{$^{2}$Brubotics, Vrije Universiteit Brussel and IMEC, Brussel, Belgium.\\
Yuya.Hamamatsu@taltech.ee}
}}

\begin{document}

\maketitle
\pagestyle{empty}

\begin{abstract}
Efficient flapping propulsion hinges on operating within a narrow Strouhal number window, a principle nature has converged upon for maximum thrust-to-power ratio. We translate this bioinspired empirical rule into real-time control, demonstrating it on an autonomous underwater vehicle driven by four soft fins. The proposed Strouhal-aware Model Predictive Control (MPC) enhances a quasi-steady hydrodynamic model with an explicit penalty for Strouhal deviation, solving the resulting nonconvex problem via a two-stage sampling and gradient optimization that runs onboard at 25 Hz. Pool and field trials show that the controller keeps each fin within the optimal Strouhal corridor (0.25-0.35) while precisely tracking commanded forces. This results in a mean reduction in mechanical power of 8.8\% to 32\% throughout the cruising range of 0.1 to 0.3 m/s. The proposed method also allows for a velocity of 0.4 m/s, which is unattainable for a baseline of the conventional inverse model. The results confirm that embedding first-principle flow physics into an MPC objective yields tangible endurance gains without sacrificing agility, offering a generic pathway to energy-aware locomotion in next-generation multifin robots.
\end{abstract}

\begin{IEEEkeywords}
Strouhal number, model predictive control, underwater robot, biomimetics, fin actuation
\end{IEEEkeywords}

\section{Introduction}
Underwater robots are increasingly turning to biology for clues on how to move quietly and efficiently in the challenging aquatic environment. For example, fish, cetaceans, and marine reptiles cruise for hours on minimal metabolic energy, relying on oscillatory lifting surfaces that exploit rather than combat the surrounding flow\cite{zhang2024review}. Replicating these flapping gaits on robotic platforms therefore promises an improvement in endurance.

A unifying metric to describe such locomotion is the dimensionless Strouhal number, which couples beat frequency, peak‑to‑peak amplitude, and forward velocity. Interestingly, a wide spectrum of swimmers and flyers, from lampreys and tunas to sea turtles and even insects, converge in $St,\approx,0.2$ -0.4 when cruising at their optimal energetic pace \cite{Triantafyllou1993,Triantafyllou1995,Taylor2003,Eloy2012,Rohr2004,OnTheRules2017,Floryan2018}.  For the swimming motion of Chelonia mydas, for example, field measurements report $St = 0.24$, mostly inside this narrow band \cite{biomimetics10050272}.  This across phyla agreement suggests that evolution has adjusted the primary kinematic factor of oscillatory propulsion to a nearly universal optimal level, maximizing hydrodynamic output per unit input.

The physical explanation behind this convergence is also established with hydrodynamic experiments and numerical simulations. At the optimal Strouhal number, the wake rolls into a coherent reverse–Kármán vortex street whose paired vortices channel a momentum jet directly aft, producing thrust with minimal wasted swirl \cite{berlinger2021fish}. Departures on either side of this band disrupt the staggered vortex pattern. In both cases, it returns to the drag-inducing wake typical of fixed foils \cite{triantafyllou1993reverse}. Therefore, maintaining a correct Strouhal window is a key requirement for locomotion optimization in robotics to achieve efficient flapping. 

Flapping propulsion has unsteady wakes that simple drag models miss and we cannot simulate at control rates. The optimal Strouhal range is a robust rule of thumb for many animals. So we use it directly in the Model Predictive Control (MPC) cost to keep Strouhal number in range, which improves efficiency in real-time compared with the model-based baseline. In addition, we propose a hybrid control scheme that balances efficiency and maneuverability. This control scheme enables the MPC to balance the need to remain within the optimal Strouhal band with the task of generating a required force and optimizing the entire flapping motion for maximum efficiency. The integration of MPC with fin-actuated robot control has been an active area of research in recent years. In particular, for the control of multiple tasks and constraints, such as the simultaneous following of paths and the avoidance of obstacles for a robotic fish \cite{wang2023trajectory}, and the control of an amphibious robot \cite{qiao2023cpg}.

Our second contribution is to extend the idea from a single-foil case to a four-fin underwater vehicle, where all flexible fins must be coordinated. Previous work shows that overall efficiency depends on each fin staying within the optimal Strouhal range, even in tandem layouts \cite{mignano2019passing,muscutt2017performance}.  By combining a body‑level force allocation module with individual per‑fin MPC loops, we allow each fin to pursue its own Strouhal optimum while still contributing coherently to the vehicle‑level force and moment demands. The design combines a body-frame controller with locomotion controllers to efficiently perform the target task.

However, real-time optimal control of autonomous underwater vehicles (AUVs) with multiple fins remains a formidable challenge. The hydrodynamics and thrust output of soft fin motion are strongly nonlinear and history dependent \cite{sun2024investigation}. Enforcing the narrow Strouhal corridor for optimal efficiency further complicates the problem, creating a highly non-convex cost surface riddled with misleading local minima \cite{wang2025optimal}. This rugged optimization landscape poses a dilemma for conventional methods: gradient-based Nonlinear MPC (NMPC) risks converging to a poor local solution. To overcome this challenge, we propose a two-stage optimization loop that synergizes the strengths of both sampling-based and gradient-based methods. In the first stage, a lightweight Sampling-based MPC (SMPC) \cite{pezzato2025sampling} generally surveys the solution space to efficiently identify a promising region of interest while avoiding the most deceptive local minima. This promising candidate then serves as a warm start for a computationally efficient NMPC that rapidly refines the solution to a precise local optimum. Hybrid or warm-start ideas of this kind have recently proven effective in underwater and other high-dimensional robotic systems \cite{caldwell2010motion, schwenkel2020online}. Our framework brings the same advantages to the generation of flapping motions with high degrees of nonlinearity and complexity, which is our case.

In summary, the contributions include the following: 
\begin{itemize}
    \item A novel MPC scheme that optimizes fin motion using a bioinspired Strouhal efficiency 
    \item A real-time hybrid optimization method for solving non-convex optimization problems by applying stepwise MPC
    \item An analysis of how Strouhal-based actuation improves energy efficiency in a flapping propulsion context using an actual robot platform
\end{itemize}

The following sections detail the development of this controller and its experimental evaluation. We show that Strouhal-guided multi-fin actuation can substantially reduce the energy of transport of an underwater robot while preserving stability and achieve high efficiency through a synergy of bioinspired empirical rule and optimal control.

\begin{figure}[t]
\includegraphics[width=1.00\linewidth]{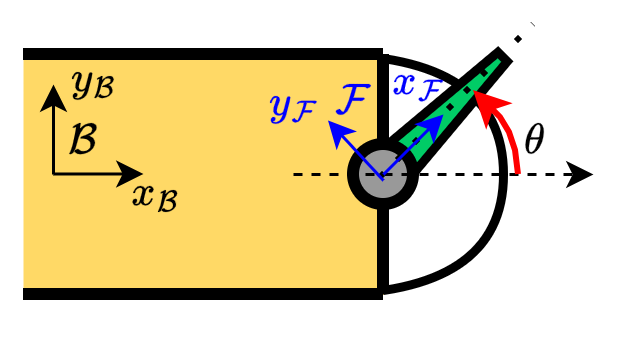}
\caption{Definition of the coordinate frames used in the proposed model. The figure illustrates the body-fixed frame {$\mathcal B$} ($x_B, y_B$) and the local fin frame {$\mathcal F$} ($x_F, y_F$). The fin's motion is described by the angle $\theta$ relative to the body frame's longitudinal axis. The hydrodynamic force vector $F$, generated by the fin, is first calculated in the local fin frame {F} and is subsequently transformed into the body frame {$\mathcal B$} for vehicle dynamic calculations.}
\label{axis}
\end{figure}

\section{Method}

In this section, we explain the fin modeling and the control based on the Strouhal number.

\subsection{Modeling}

We explain the modeling method for the fins of a fin-actuated robot. Fig. \ref{axis} shows the configuration of the axis of the robot and the fins.

Let $\{\mathcal B\}=\{x_{\mathcal B},y_{\mathcal B},z_{\mathcal B}\}$ denote the fixed body frame and $\{\mathcal W\}$ the inertial frame. The frames are related by the Euler angle rotation $(\phi_b,\theta_b,\psi_b)$.  Due to the dominant forces of the fin that act in the cross-sectional plane $x_{\mathcal B}\!-\!y_{\mathcal B}$, we model each fin using a single tip deflection angle $\theta\in[-\pi,\pi]$, which yields a two-state planar description.

One flapping cycle is divided into $H$ intervals indexed by $k\in\{1,\dots,H\}$. With start and end angles $\theta_{k}$ and $\theta_{k+1}$, the stroke angle and mean angular rate per interval are

\begin{equation}
\begin{aligned}
d_k=\theta_{k+1}-\theta_{k}, \qquad
\omega_k=\frac{d_k}{\Delta t_k},
\end{aligned}
\end{equation}
where $\Delta t_k$ is the duration of the interval selected by the MPC. For a fin with a radius of gyration around the pivot axis $r_c$, the tip velocity is

\begin{equation}
v_k=r_c\,|\omega_k|.
\end{equation}

\begin{figure}[t]
\includegraphics[width=1.00\linewidth]{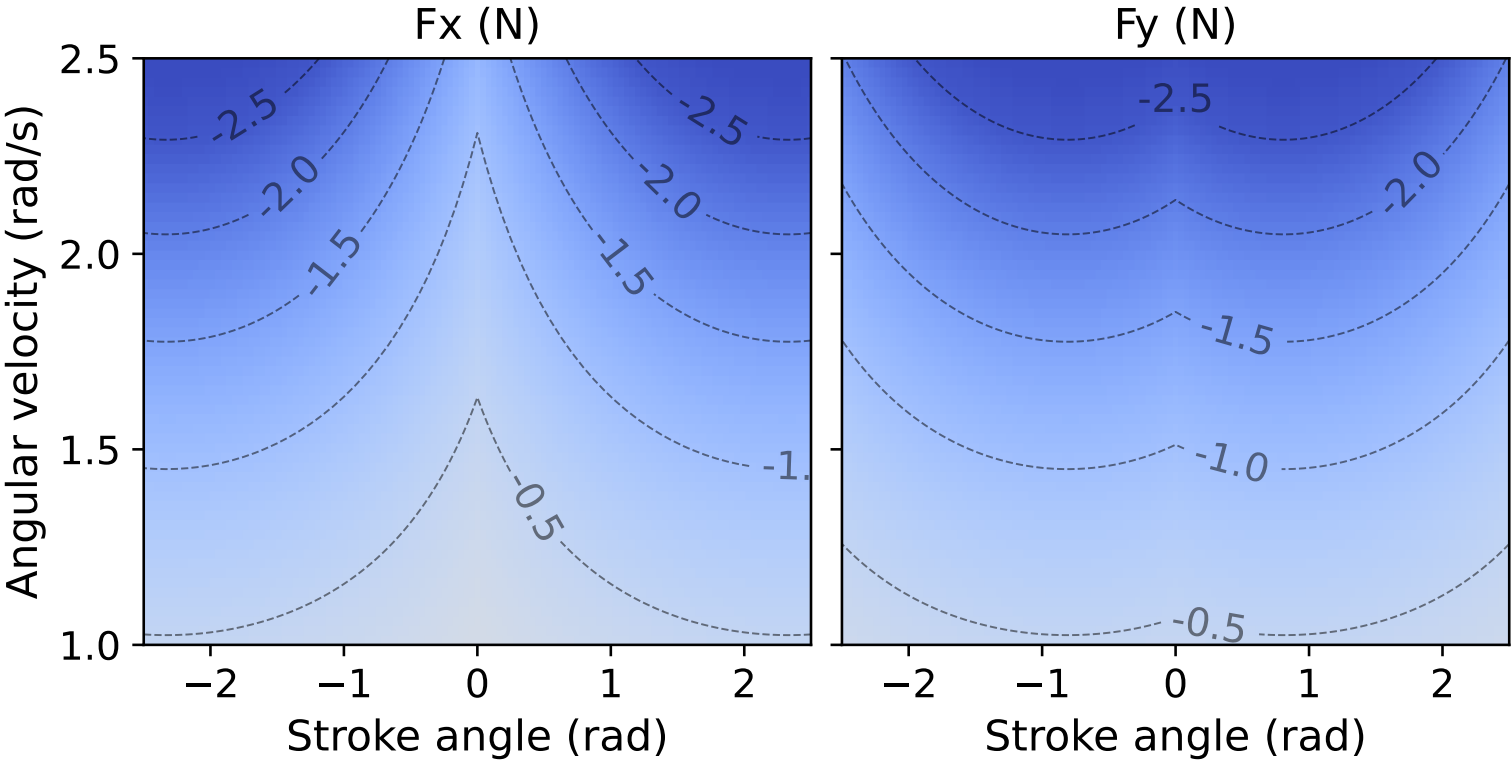}
\caption{The visualizations of the generated force in the local fin frame $\{\mathcal F\}$.}
\label{forcemap}
\end{figure}

Assuming that the boundary layer development is fast compared with the stroke period, we employ a quasi-steady approximation of the hydrodynamic modeling of fins. This modeling choice represents prioritizing the computational tractability required for real-time MPC over a complete representation of unsteady fluid dynamics \cite{kopman2014dynamic}. Using the lift and drag coefficients $C_L$ and $C_D$ identified from the available experimental data \cite{hamamatsu2025underwater}, the drag $D_k$ and the lift $L_k$ in the interval $k$ are

\begin{equation}
\begin{aligned}
D_k=\tfrac12\rho C_D A_f v_k^{2}, \qquad
L_k=\tfrac12\rho C_L A_f v_k^{2},
\end{aligned}
\end{equation}
where $\rho$ is the density of water and $A_f$ the area of the fin planform. Because the angle of attack varies with stroke angle, the force direction is approximated by the half-opening angle

\begin{equation}
\theta_{r,k}=\frac{d_k}{2}.
\end{equation}

Expressed in the local fin frame $\{\mathcal F\}$ as shown in Fig. \ref{axis}, the mean force vector $\mathbf f_k^l=(f_{x,k}^{\,l},f_{y,k}^{\,l})^{\top}$ becomes

\begin{equation}
\begin{aligned}
f_{x,k}^{\,l} &= -\,L_k\cos\theta_{r,k}-D_k\sin\theta_{r,k}, \\
f_{y,k}^{\,l} &= -\,L_k\sin\theta_{r,k}-D_k\cos\theta_{r,k}.\
\end{aligned}
\end{equation}

\begin{figure}[t]
\includegraphics[width=1.00\linewidth]{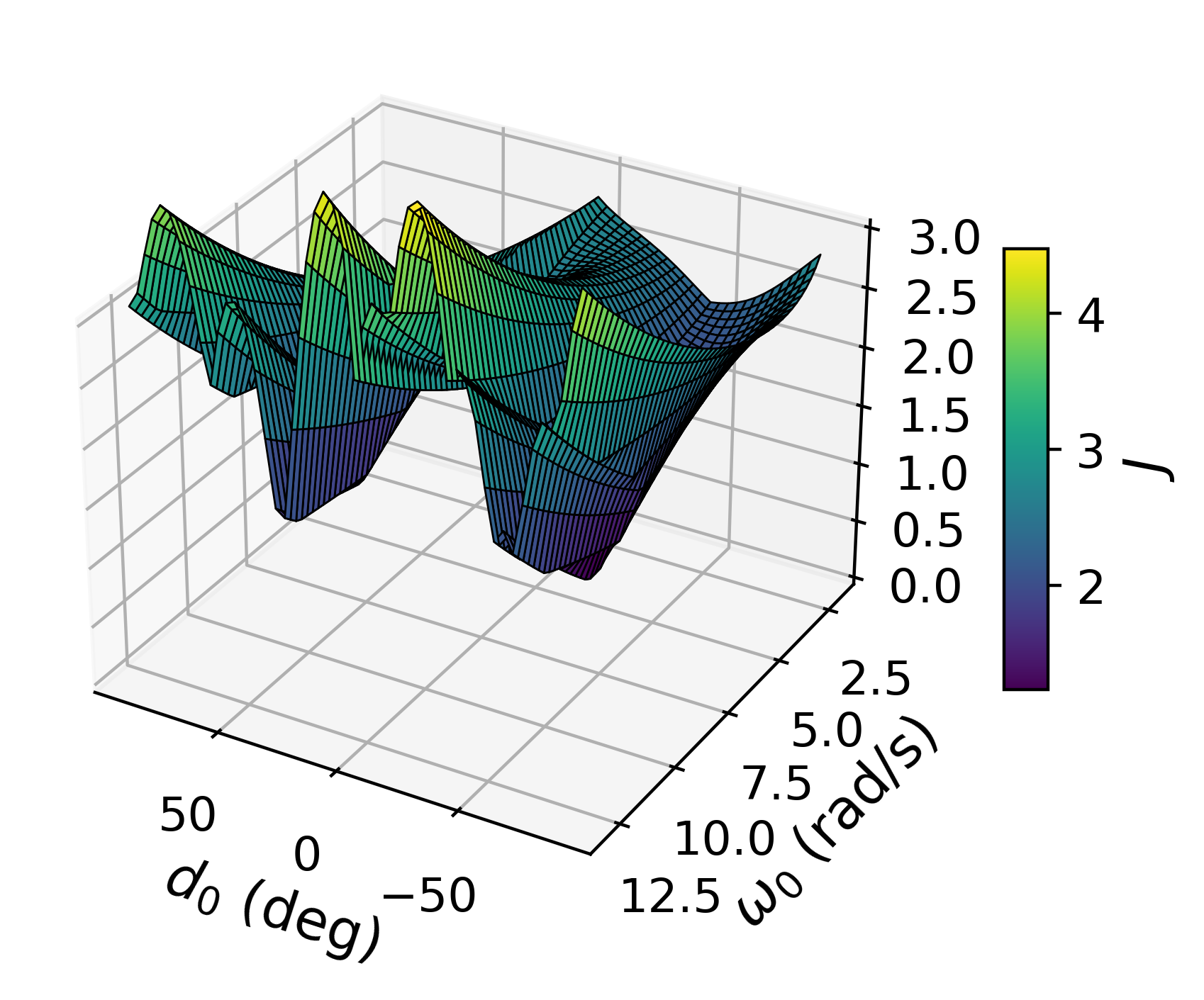}
\caption{
The plot shows $J(d_0,d_1,\omega_0,\omega_1)$ after freezing the second interval at $d_1=-40^{\circ}$ and $\omega_1=3~\text{rad/s}$.  The two displayed axes sweep the first-interval stroke angle, color and height encode the resulting cost $J$. Other parameters follow those of Table I with a desired body-frame thrust $(-0.5,\,0.5)\,\text{N}$ and a forward velocity is set as $U=0.3~\text{m/s}$. 
}
\label{cost}
\end{figure}

Fig. \ref{forcemap} shows the force generated in the local fin frame $\{\mathcal F\}$. Transformation into the body frame $\{\mathcal B\}$ uses mid-stroke angle

\begin{equation}
\theta_{m,k}=\theta_k+\frac{d_k}{2},
\end{equation}
and the two-dimensional rotation matrix,

\begin{equation}
\mathbf R(\theta_{m,k})=
\begin{bmatrix}
\cos\theta_{m,k}&-\sin\theta_{m,k}\\
\sin\theta_{m,k}& \phantom{-}\cos\theta_{m,k}
\end{bmatrix}.
\end{equation}
Hence,

\begin{equation}
\mathbf f_k=\mathbf R(\theta_{m,k})\,\mathbf f_k^l
=\begin{bmatrix}f_{x,k}\\f_{y,k}\end{bmatrix}_{\!\mathcal B}.
\end{equation}

Equations (3)-(8) show that $\mathbf f_k$ depends not only on $(d_k,\omega_k)$ but also on the preceding angle $\theta_k$. To accelerate MPC, the space $(d,\omega)$ is pre-tessellated in a uniform grid; for each node, $(f_x^l,f_y^l)$ is tabulated by (3) - (5) and later obtained by bilinear interpolation during optimization. 

\begin{figure*}[t]
\includegraphics[clip, width=18cm]{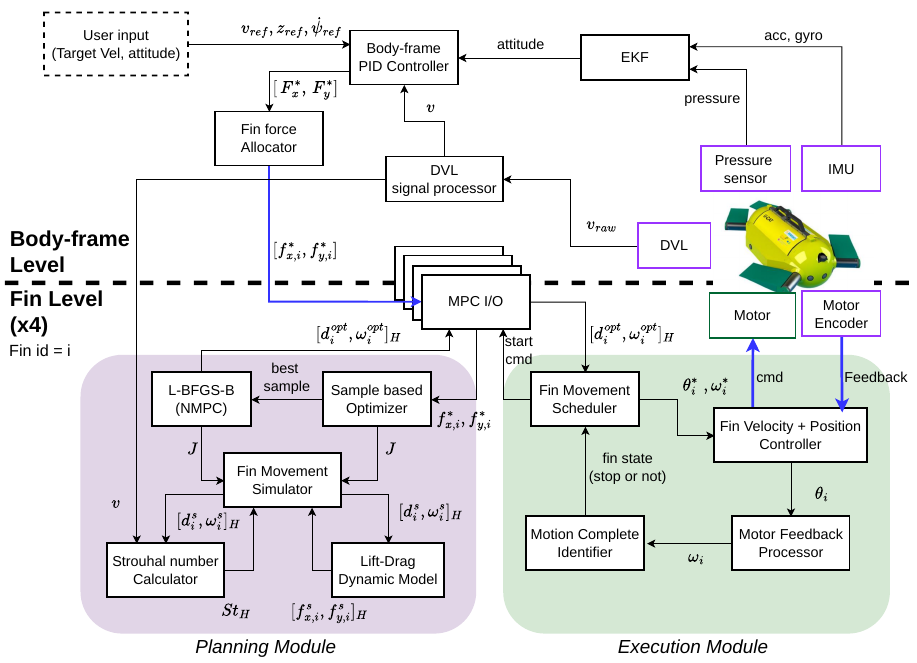}
\caption{System architecture of the proposed Strouhal-aware control framework. The Body-frame Level controller generates high-level thrust commands based on velocity and attitude feedback. These commands are distributed to four parallel Fin Level controllers. They consists of two components. First one is Fin MPC module that optimizes the motion to track the commanded force while adhering to Strouhal number constraints. Second one is Fin Servo Manager that executes the planned trajectory by MPC module to the fin.}
\label{diagram}
\end{figure*}

\subsection{Fin-level Model Predictive Control Architecture}

At the beginning of every control cycle, the controller computes a short stroke sequence by solving a finite-horizon optimization. The objective function balances the tracking error of the commanded mean thrust vector in the body frame with hydrodynamic efficiency, as well as motion smoothness and adherence to actuation and rate limits. The four fins are planned independently, and each fin executes a control cycle that includes multiple half-strokes represented by $H$ consecutive angular displacements. 
Let $\theta_k$ and $\theta_{k+1}$ be the start and end angles of the $k$-th sub-interval, with $k\in\{1 .. H\}$.
The output variables are collected in vector.

\begin{equation}
\begin{aligned}
\boldsymbol{u}
\;=\;
\bigl[d_1,d_2,..., d_H, \;\omega_1,\omega_2, ... , \omega_H\bigr]^{\!\top}\in\mathbb R^{2H}
\end{aligned}
\end{equation}

 The displacement $d_k$ can assume either sign; its magnitude is limited by the mechanical range $d_{\max}=A_{\max}/L_f$, while the angular velocity is bounded by the actuator limits $\omega_{\min}\le\omega_k\le\omega_{\max}$. \RED{We set $H=2$ so that the optimizer spans one full flapping cycle; 
$H=1$ is found to make periodic motion harder to induce, while larger horizons do not produce a consistent overall improvement.} For these locomotion procedures, the Strouhal number is calculated as

\begin{table}[t]
\centering
\caption{Key parameters used in the system.}
\begin{tabular}{lll}
\toprule
\textbf{Parameter} & \textbf{Symbol / meaning} & \textbf{Value} \\
\midrule
\multicolumn{3}{c}{\textit{MPC parameters}}\\
Weights                            & $\lambda_f, \lambda_{St}, \lambda_a, \lambda_d$ & 1.5, 5.0, 1.0, 1.0 \\
\RED{Horizon length}                 & $H$                            & 2 \\
Sampling candidates                & $N$                            & 500 \\
NMPC max iterations                & $k_{\max}$                     & 60 \\
Fluid density                      & $\rho$                         & 1000\,kg\,m$^{-3}$ \\
Fin planform area                  & $A_f$                          & 0.020\,m$^{2}$ \\
Effective fin length                  & $L_f$                          & 0.10\,m \\
Radial coefficient                 & $r_c$                          & 0.10\,m \\
Optimal Strouhal number             & $St_{opt}$                     & 0.30 \\
Drag coefficient                   & $C_D$                          & 0.70 \\
Lift coefficient                   & $C_L$                          & 0.25 \\
\midrule
\multicolumn{3}{c}{\textit{Body-frame PID parameters}}\\
Velocity PID gains                 & $K_P,\,K_I,\,K_D$              & 40.0,\;1.0,\;0.1 \\
Velocity actuator limit            & -                     & 2.5\,N \\
Depth PID gains                    & $K_P^{d},\,K_I^{d},\,K_D^{d}$  & 5.0,\;0.1,\;0.005\\
Depth actuator limit               & -                & 0.5\,N \\
Yaw PID gains                    & $K_P^{y},\,K_I^{y},\,K_D^{y}$  & 3.0,\;0.2,\;0.0\\
Yaw actuator limit               & -                 & 0.5\,N \\
\bottomrule
\end{tabular}
\end{table}

\begin{table}[t]
    \caption{Robot specification}
    \label{tab:robotparam}
    \begin{center}
    \begin{tabular}{c c}

    \textbf{Name} & \textbf{Value} \\
    \hline
    Weight & 22.0 kg \\
    Computer & Jetson Orin Nano \\
    Motors & Maxon Motor 272763 \\
    Motor controller & Maxon Motor EPOS2 \\
    IMU & Microstrain 3DM-CX5-IMU  \\
    Pressure sensor & Gems 3101P \\
    DVL & Waterlink A50 \\
    Fin material & Zhermack Elite Double 22 \\ 
    \hline
    OS & Ubuntu 22.04 \\
    Middleware & ROS 2 Humble \\
    \hline
    \end{tabular}
    \end{center}
\end{table}

\begin{equation}
\begin{aligned}
St
=
\frac{2 f A}{V_B}, \qquad
f \;=\;\frac{H}{2\sum_{k=1}^{H}\!\Delta t_k},
\\
A \;=\;\frac{1}{H}\sum_{k=1}^{H}\!A_k, \qquad
A_k=\frac{|d_k|L_f}{2}.
\end{aligned}
\end{equation}

They relate the fin beat frequency $f$, the mean peak-to-peak amplitude $A$, and the swimming velocity $V_B$. Extensive biological observations and unsteady-flow experiments agree that the propulsive efficiency of flapping foils peaks in a narrow corridor, $0.25\lesssim St\lesssim0.35$. However, the penalty for deviating from this optimum is not symmetric. According to the model proposed by Floryan et al. \cite{Floryan2018}, the drop in efficiency is more gradual for Strouhal numbers above the peak compared to those below it. This suggests that operating at slightly too high a frequency is less detrimental than flapping too slowly, as the latter can result in an inability to form a cohesive, thrust-producing vortex street. To incorporate this nuanced asymmetric characteristic into the controller, we introduce an efficiency surrogate as follows:

\begin{equation}
\begin{aligned}
\eta_{\text{hat}}(St)
\;=\;
\eta_{\max}\,
\frac{St}{St_{\text{opt}}}
\,\exp\!\bigl[1-St/St_{\text{opt}}\bigr], \\
\end{aligned}
\end{equation}
which attains its maximum $\eta_{\max}$ at $St_{\text{opt}}$ and decays smoothly on either side.  A penalty term

\begin{equation}
c_{St}(St)
\;=\;
\max\!\bigl(0,\;\eta_{\max}-\eta_{\text{hat}}(St)\bigr),
\end{equation}
is therefore zero in the optimal corridor and grows proportionally to the expected loss in propulsive efficiency elsewhere.

The mean thrust commanded on the body frame is denoted $\mathbf{f}^{*}=(f_x^{*},f_y^{*})^{\!\top}$.
With stroke-wise forces $\mathbf{f}_k(d_k,\omega_k;\theta_k)$, the time-weighted cycle average becomes

\begin{equation}
\begin{aligned}
\bar{\mathbf{f}}
\;=\;
\sum_{k=1}^{H} W_k\,\mathbf{f}_k , \qquad
W_{k}=\frac{\Delta t_k}{\sum_{k=1}^{H}\Delta t_k} , \qquad
\Delta t_k=\frac{|d_k|}{|\omega_k|}. 
\end{aligned}
\end{equation}

The tracking error is measured in the one-norm,

\begin{equation}
c_{\text{force}}(\boldsymbol{u})
\;=\;
\bigl|\,\bar f_x-f_x^{*}\bigr|
\;+\;
\bigl|\,\bar f_y-f_y^{*}\bigr|.
\end{equation}

The reference force $(f_x^{*},f_y^{*})^{\!\top}$ is assigned based on the configuration of the fins on the body frame. The implementation of the assignment in the velocity control mechanism of the body frame will be explained in Section 3. It is also known that within the optimal Strouhal number's band, larger amplitudes tend to improve efficiency \cite{ding2024effect}, \cite{liou2024strouhal}. Therefore, the following penalty, which activates only below a prescribed minimum amplitude $A_{\min}$,

\begin{equation}
c_{\text{amp}}(\boldsymbol{u})
\;=\;
\sum_{k=1}^{H}\max\!\bigl(0,\;A_{\min}-A_k\bigr).
\end{equation}

To avoid abrupt switching between consecutive control cycles, we add a temporal-regularization term that biases the optimizer toward the previously executed solution $\boldsymbol{u}^{\text{prev}}$  \cite{Eloy2012}. This inertia discourages chattering between nearly-equivalent stroke patterns, promotes continuity of fin angles and rates across cycle boundaries \cite{Eloy2012}.

\begin{equation}
c_{\text{diff}}(\boldsymbol{u})
\;=\;
\frac12
\sum_{k=1}^{H}\!
\bigl(\,
|d_k-d_{k}^{\text{prev}}|
+|\omega_k-\omega_{k}^{\text{prev}}|
\bigr).
\end{equation}

The multi-objective cost reads

\begin{figure*}[t]
\includegraphics[clip, width=18cm]{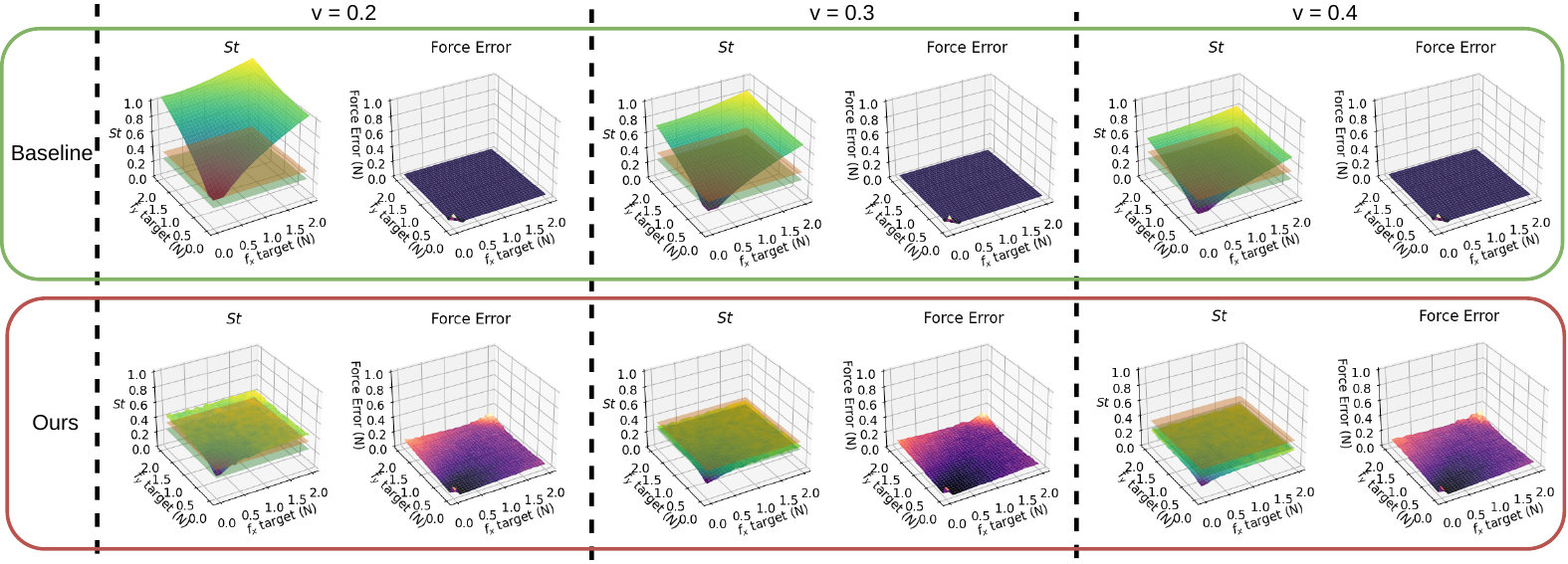}
\caption{Strouhal–error landscape for the baseline inverse model and the proposed controller.
For each forward velocity examined ($v_b = 0.20,\; 0.30,\; 0.40\;\text{m/s}$; columns separated by dashed vertical lines), a pair of 3-D surfaces is displayed: the left-hand surface shows the single-fin Strouhal number $St$ as a function of the target force components $(f_x,f_y)$ and the right-hand surface shows the total force-magnitude error $|\Delta F|$ obtained after allocating the command to the four fins and re-evaluating the hydrodynamic model.  The translucent beige plane marks the efficiency optimum at $St = 0.30$.  In the baseline (green frame, top row) the analytical inverse model of Remmas et al. is used directly; $St$ rises almost linearly with the requested force and rapidly leaves the efficient band.}
\label{stanalysis}
\end{figure*}

\begin{equation}
J(\boldsymbol{u})
=
\lambda_f\,c_{\text{force}}
+\lambda_{St}\,c_{St}
+\lambda_a\,c_{\text{amp}}
+\lambda_d\,c_{\text{diff}},
\end{equation}
subject to the box constraints,

\begin{equation}
\begin{aligned}
|d_k|\le d_{\max} ,\qquad \omega_{\min}\le\omega_k\le\omega_{\max}
\end{aligned}.
\end{equation}

\begin{algorithm}[t]
\caption{Planning Module}
\label{alg:planning_module}
\begin{algorithmic}[1]
\REQUIRE Target thrust $\boldsymbol{f}_{\text{target}}$, body velocity $v$, previous angle $\theta_{\text{prev}}$, previous output $\boldsymbol{u}_{\text{prev}}$
\ENSURE Command array $\boldsymbol{u}_{\text{cmd}}$
\STATE \textbf{Stage 1: Random Sampling}
\FOR{$i = 1$ \textbf{to} $N$}
    \STATE $\boldsymbol{u}_i \leftarrow \textsc{RandomStrokePair}()$ \COMMENT{output vector $\boldsymbol{u}_i$} 
    \STATE $\bar{\boldsymbol{f}}_i \leftarrow \textsc{MeanThrust}(\boldsymbol{u}_i, \theta_{\text{prev}})$
    \STATE $\text{St}_i \leftarrow \textsc{Strouhal}(\boldsymbol{u}_i, v)$
    \STATE $J_i \leftarrow \text{CalculateCost}(\bar{\boldsymbol{f}}_i, \text{St}_i, ...)$
\ENDFOR
\STATE $\boldsymbol{u}^{\text{s}} \leftarrow \displaystyle\arg\min_{\boldsymbol{u}_i} J_i$ \COMMENT{Find the best sample}
\STATE $\boldsymbol{u}^{\text{SMPC}} \leftarrow \boldsymbol{u}^{\text{s}}$ \COMMENT{Set as warm start for NMPC}
\vspace{0.5em}
\STATE \textbf{Stage 2: Local NMPC Refinement}
\STATE $\boldsymbol{u}^{*} \leftarrow \text{L-BFGS-B}(J, \boldsymbol{u}^{\text{SMPC}}, \text{bounds})$ \COMMENT{Optimized output vector $\boldsymbol{u}^*=[d_1^*,...,d_H^*, \omega_1^*,...,\omega_H^*]$}
\STATE $\theta_0^* \leftarrow \theta_{\text{prev}}$
\FOR{$k = 1$ \textbf{to} $H$}
    \STATE $\theta_k^* \leftarrow \theta_{k-1}^* + d_k^*$ \COMMENT{Calculate absolute target angles}
\ENDFOR
\STATE $\boldsymbol{u}_{\text{cmd}} \leftarrow [\theta_1^*, \omega_1^*, ..., \theta_H^*, \omega_H^*]$ \COMMENT{Make command sequence}
\RETURN $\boldsymbol{u}_{\text{cmd}}$
\end{algorithmic}
\end{algorithm}

The cost surface in the $2H$-dimensional output space is nonconvex due to the sign constraints and the piecewise definitions (17) and (18), as shown in Fig. \ref{cost}.  Pronounced ridges and multiple isolated minima illustrate the strong nonconvexity that motivates the hybrid SMPC + NMPC solver.

In the first stage, the controller draws $N$ random stroke pairs respecting the limits (18).  
The optimization routine is organized as a two-stage sequence that cleanly separates broad exploration from rapid local refinement while respecting the real-time constraints of the controller. At the start of each control interval, a sampler draws a finite set of admissible control sequences that already satisfy the box constraints of (18).  Each candidate is propagated through the predictive model described in Section II.A, and its performance is evaluated by the cost function.  Because this evaluation requires only table look-ups and elementary arithmetic, it executes quickly even when the sample set is large.  The candidate with the lowest cost, denoted $\boldsymbol{u}^{\mathrm{SMPC}}$, serves as a feasible, near-global warm start for the next stage. 

The warm start $\boldsymbol{u}^{\mathrm{SMPC}}$ is passed to the bound-constrained nonlinear model predictive controller.  We solve this problem with the simple bound limited memory Broyden–Fletcher–Goldfarb–Shanno (BFGS) algorithm (L-BFGS-B) \cite{zhu1997algorithm}, which preserves feasibility by projecting every step of the test back onto the constraint box. Gradients are obtained by forward finite differences sufficient given the modest dimensionality of the output vector, but analytic or algorithmic differentiation could be substituted when higher accuracy or larger problems make that advantageous.  The solver converges well within the available control budget, providing a refined solution $\boldsymbol{u}^{\star}$ that reduces the objective value. The two-stage procedure guarantees that the final solution satisfies all hard bounds by construction, while the Strouhal penalty in the cost functional keeps feasible motions close to the empirically optimal regime.  By combining a coarse global search with a fast gradient-based polish, the controller avoids poor local minima.  The same structure can be transferred to the low-level motor controller of another fin by replacing the model, the cost terms, or the limits of the actuator without altering the optimization flow itself.

\begin{algorithm}[t]
\caption{Execution Module}
\label{alg:execution_module}
\begin{algorithmic}[1]
\REQUIRE Command sequence $\boldsymbol{u}\in\mathbb R^{2H}$, feedback $(\theta_{\text{fb}}, \omega_{\text{fb}})$
\ENSURE Motor references $(\theta_{\text{ref}}, \omega_{\text{ref}})$

\IF{current interval finished}
    \STATE $(\theta_{\text{ref}}, \omega_{\text{ref}}) \leftarrow \textsc{PopNext}(\boldsymbol{u})$
    \STATE \textsc{SendToServo}$(\theta_{\text{ref}}, \omega_{\text{ref}})$
    \STATE $\textit{timer\_start} \leftarrow \textsc{now}()$
    \STATE current interval finished $\leftarrow$ \textbf{False}
\ENDIF
\vspace{0.5em}

\STATE $\delta\omega \leftarrow \omega_{\text{fb}} - \omega_{\text{prev}}$
\STATE $a \leftarrow \delta\omega / \Delta t_{\text{fb}}$
\STATE $\omega_{\text{pred}} \leftarrow \omega_{\text{fb}} + a\!\cdot\!\bigl(\textsc{now}()-t_{\text{fb}}\bigr)$
\IF {$|\omega_{\text{pred}}| < \omega_{\text{th}}$}
    \STATE current interval finished $\leftarrow$ \textbf{True}
\ENDIF
\vspace{0.5em}

\IF{$\boldsymbol{u}$ is empty \textbf{and} current interval finished}
    \STATE \textsc{SendCompletionSignal}() \COMMENT{Triggers next cycle}
\ENDIF
\end{algorithmic}
\end{algorithm}

\subsection{Baseline controller for comparison}

\RED{As a practical non-optimizing reference, we employ the inverse model-based controller \cite{remmas2021inverse}, which is representative of the fixed-frequency, amplitude-modulated control commonly used in fin-actuated AUVs \cite{ding2024experimental}. Each fin is actuated at a fixed frequency $f$, and the required mean fin force $\overline{f^*}$ is directly mapped to the maximum stroke angle by an empirical relation.}

\begin{equation}
\theta_{\text{max}} \;=\; 
\arccos\left(
      \frac{-\,\overline{f^*}}
           {8\,C_d\,\rho\,A_f \bigl(r_c \pi f\bigr)^2}
      + 1
      \right),
\end{equation}
where $F\,[\mathrm N]$ is the thrust or lift assigned to that fin. In this model, the two-dimensional force assigned $(f_x^{*},f_y^{*})^{\!\top}$ is output by switching the vibration center (zero direction).

\section{Implementation}

In this section, we explain the architecture and algorithms for implementing the techniques described in the previous section on an actual robot. Fig. \ref{diagram} shows the whole robot system diagram.

\subsection{Hardware configuration}
This study was carried out using the U-CAT \cite{salumae2014design}, an underwater robot propelled by four soft-fin actuators. This AUV was equipped with a pressure sensor for precise depth measurement and an inertial measurement unit (IMU) for attitude estimation. The experimental setup also included a Doppler Velocity Log (DVL) mounted at the rear of the vehicle to acquire velocity data. Before each experiment, the AUV was adjusted to achieve neutral buoyancy. The hardware configuration of the robot is summarized in Table \ref{tab:robotparam}.

\begin{table*}[t]
\centering
\caption{\RED{Ablation Study on Overall Performance}}
\label{tab:ablation_study}
\begin{tabular}{lcccc}
\hline 
\textbf{Configuration} & \textbf{Force Error [N]} & \textbf{St Deviation (from 0.30)} & \textbf{In-Range Rate [\%]}  & \textbf{Command Stability} \\ \hline
Ours (Full MPC) & 0.1559 & 0.0285 & 88.0 &  1.625 \\
w/o Strouhal penalty & 0.1466 & 0.1227 & 34.7 & 1.761 \\
w/o Amplitude penalty & 0.1541 & 0.0320 & 86.0 & 1.801 \\
w/o Strouhal and Amplitude penalties & 0.1323 & 0.1259 & 30.0 & 1.981 \\
w/o smoothness penalty & 0.1488 & 0.0261 & 89.3 & 2.177 \\  \hline 
\end{tabular}
\end{table*}

\subsection{Fin force allocation and body frame control}

In order to generate the proposed fin motion, it is necessary to assign force from the body frame to the fins attached to the robot. In this study, we implement a velocity control mechanism that maintains a constant whole-body velocity and depth of the body frame. The body frame layer receives the desired forward velocity and dive depth from the navigation stack and converts them into a mean thrust vector of the body frame
$\mathbf{F}^{*}= [\,F_x^{*},\,F_y^{*}]^{\!\top}$.
That vector is in turn forwarded to the described MPC.  The forward velocity is measured by the DVL and the depth by a pressure sensor; both signals are first smoothed by a short median filter to suppress outliers and then regulated by independent PID controllers.  The velocity loop produces the required axial thrust $f_x^{*}$, while the depth loop modulates the vehicle’s pitch rather than generating net vertical force: a positive depth error is corrected by commanding the front fins to generate a downward lift and vise versa. Similarly, the yaw rate controller uses feedback from the IMU's gyroscope to generate a corrective yaw moment. This moment is realized by creating a thrust differential between the two front fins, causing one to produce more forward thrust than the other to initiate a turn. Consequently, the thrust allocation proceeds as follows.  The axial component is distributed uniformly,

\begin{equation}
    f_{x,i}^{*}= \tfrac14\,F_x^{*}, \qquad i = 0,\dots,3
\end{equation}
so that all four fins contribute equally to propulsion.
The vertical component arising from the depth controller is applied only to the two front fins,

\begin{equation}
    \begin{aligned}
        f_{y,0}^{*}=f_{y,1}^{*}= \tfrac12\,F_y^{*} .\qquad
f_{y,2}^{*}=f_{y,3}^{*}= 0 
    \end{aligned}
\end{equation}

The force assigned to each fin as described above is transferred to the module described in the next section.

\begin{figure*}[t]
\includegraphics[width=18cm]{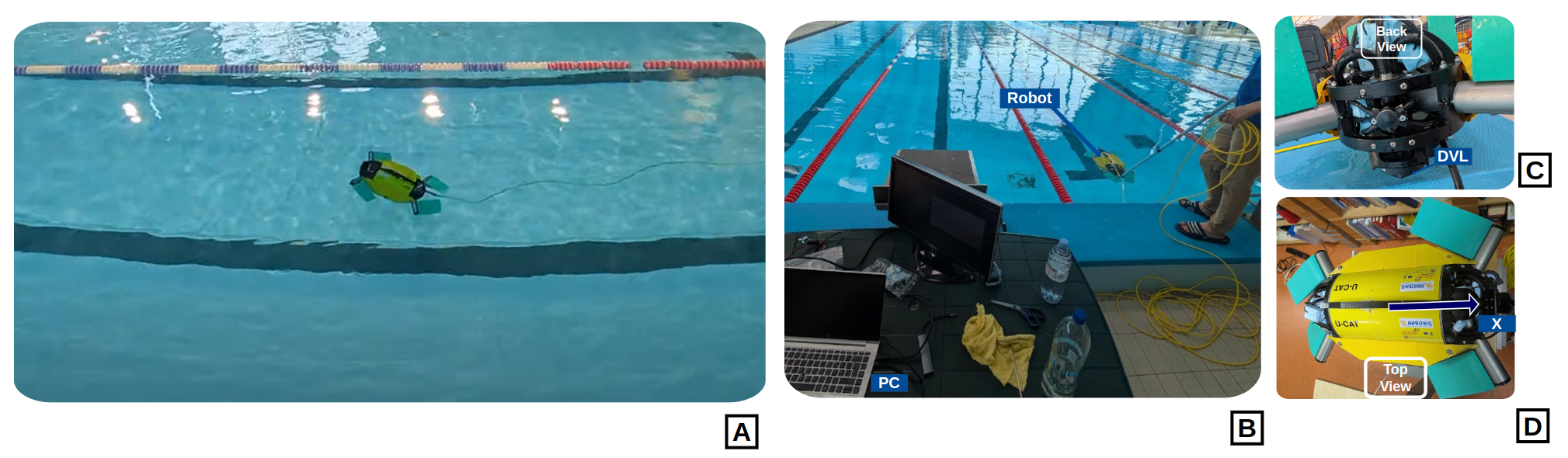}
\caption{Experimental setup for validating the Strouhal-aware MPC framework. (A) The U-CAT robot, equipped with four flexible fins, performs a trajectory in the test pool. (B) A laptop PC at the poolside serves as the control and data acquisition station. (C) A Doppler Velocity Log (DVL) is mounted on the vehicle's rear to provide ground-truth velocity data for performance evaluation. (D) A top-down picture of the robot.}
\label{setup}
\end{figure*}

\subsection{Motion control procedure}
MPC framework proposed in this study requires sophisticated computation to derive optimal fin motions. To enable its real-time execution on the physical robot, we designed a two-layer control architecture that decouples the computationally intensive Planning Module (highlighted purple in Fig. \ref{diagram} and Algorithm 1) process from the Execution Module (highlighted with green in Fig. \ref{diagram} and Algorithm 2) process responsible for precise motor actuation. This architecture consists of a Planning Module and an Execution Module for each fin, operating independently in parallel and communicating asynchronously through sparse signal passing.

The Planning Module serves as a high-level controller that runs the MPC algorithm. It receives the target thrust setpoint and the latest robot state from the higher-level systems, then computes a sequence of fin strokes that maximizes energy efficiency, and publishes this command sequence. 
The Execution Module functions as a low-level controller that translates the command sequence $u_{cmd}=[\theta_{1}^{*},\omega_{1}^{*},...,\theta_{H}^{*},\omega_{H}^{*}]$ from the Planning Module into motor actions, dedicating itself to faithfully executing the received motion sequence. Each stroke motion is achieved through precise position and velocity control executed by a motor driver. During execution, this module continuously monitors the feedback of the angular velocity $\omega_{fb}$ from the encoder and estimates the acceleration $a$ from its first difference. Then it uses linear extrapolation to predict the future angular velocity $\omega_{pred}$ and determine if a stroke has been mechanically completed by checking if $|\omega_{pred}|$ is below a predefined threshold $\omega_{th}$. Upon detecting completion, it immediately sequences the next stroke, and once the entire sequence is exhausted, it returns a completion flag to the Planning Module. This completion signal triggers the Planning Module to calculate the next optimal stroke sequence for the subsequent interval.

This decoupled architecture is important to our work. Integrates data arriving on three distinct timescales: thrust commands and DVL velocity from the navigation stack, plus fin encoder feedback. By allowing the Planning Module to perform calculations while the Execution Module manages fin actuation, the timing consistency of the fins is preserved even under abrupt changes in thrust demand or network jitter. Consequently, this robust system allows energy efficiency optimization to run continuously without sacrificing the necessary computational margin.

\begin{table}[t]
  \centering
  \caption{Benchmark of MPC}
  \begin{tabular}{@{}r|rr|rr@{}}
    \multirow{2}{*}{Candidates $N$}
      & \multicolumn{2}{c|}{\textbf{SMPC only}}
      & \multicolumn{2}{c}{\textbf{SMPC + NMPC}}\\[-3pt]
    & Avg.\ Cost & Time [ms] & Avg.\ Cost & Time [ms] \\ \hline
     250  & 2.1690 & 0.36 & 1.8697 & 38.58 \\
     500  & 2.0948 & 0.61 & 1.8347 & 37.89 \\
   1\,000 & 2.0525 & 0.98 & 1.8340 & 39.93 \\
   1\,500 & 2.0112 & 1.14 & 1.8327 & 37.99 \\
   2\,000 & 1.9833 & 1.29 & 1.7909 & 39.93 \\
   5\,000 & 1.9899 & 2.87 & 1.7889 & 37.55 \\
  \end{tabular}
  \label{tab:mpc_benchmark_updated}
\end{table}

\begin{table*}[t]
\centering
\begin{threeparttable}
\caption{\RED{RESULT COMPARISON}}
\label{tab:energy_consumption}
\begin{tabular}{lccccccccc}
\hline
 & \multicolumn{3}{c}{\textbf{Baseline}} & \multicolumn{3}{c}{\textbf{MPC}} & \multicolumn{3}{c}{\textbf{MPC (sync)}} \\ \cline{2-10} 
\begin{tabular}{@{}l@{}}\textbf{Body Vel.} \\ {[m/s]}\end{tabular} & \begin{tabular}{@{}c@{}}\textbf{Power [W]} \\ \textbf{(Sav. [\%])}\end{tabular} & \textbf{Vel. MAE} & \textbf{St MAE} & \begin{tabular}{@{}c@{}}\textbf{Power [W]} \\ \textbf{(Sav. [\%])}\end{tabular} & \textbf{Vel. MAE} & \textbf{St MAE} & \begin{tabular}{@{}c@{}}\textbf{Power [W]} \\ \textbf{(Sav. [\%])}\end{tabular} & \textbf{Vel. MAE} & \textbf{St MAE} \\ \hline
0.1 & 1.900 (0.0) & 0.0446 & 0.3632 & 1.612 (15.1) & 0.0465 & 0.2402 & 1.732 (8.8)  & 0.0486 & 0.2626 \\
0.2 & 3.251 (0.0) & 0.0355 & 0.9507 & 2.212 (32.0) & 0.0321 & 0.1918 & 2.262 (30.4) & 0.0426 & 0.2349 \\
0.3 & 3.724 (0.0) & 0.0234 & 0.3780 & 3.049 (18.1) & 0.0389 & 0.0558 & 2.686 (27.9) & 0.0370 & 0.0544 \\
0.4 & -\tnote{a} & -\tnote{a} & -\tnote{a} & 3.341 (N/A\tnote{b}~) & 0.0754 & 0.0247 & 3.386 (N/A\tnote{b}~) & 0.0654 & 0.0370 \\ 
\hline
\end{tabular}
\begin{tablenotes}
    \item[a] Baseline controller could not achieve this velocity due to thrust limitations.
    \item[b] Because the baseline could not reach the reference velocity.
\end{tablenotes}
\end{threeparttable}
\label{tab:avg-power-eff}
\end{table*}

\subsection{Fin synchronizing}
The controller has an optional rear fin synchronization mode, which mirrors the motion of the rear two fins. With this mode active, the optimization problem is solved for three logical actuators instead of four. The commands determined by the MPC are then transferred to the left and right fin's Execution modules, which synchronize the movements of the left and right fins by matching the timing of the start of the motion.

\section{Results}
This section presents the performance of the proposed control framework, validated through numerical benchmarks and experimental pool and field trials.

\subsection{Numerical benchmarks}
This section presents a preliminary analysis of the performance of the proposed control framework, focusing on key metrics evaluated through simulation and benchmarking. These results validate the core components of the controller using robot hardware prior to full experimental trials. The capacity of the proposed MPC scheme to maintain the Strouhal number, a key indicator of locomotor efficiency, within its optimal range was evaluated. Fig. \ref{stanalysis} illustrates a comparative analysis between the proposed method and a baseline controller at three distinct forward velocities (v = 0.2, 0.3 and 0.4 m / s). The baseline controller exhibits a near-linear increase in the Strouhal number with rising thrust commands, causing a rapid deviation from the optimal efficiency band. In contrast, the proposed MPC framework successfully regulates the Strouhal number, confining it within the optimal range of 0.25 to 0.35 throughout the operational envelope. This is accomplished while simultaneously maintaining a precise thrust tracking error of less than 0.2 N. This outcome demonstrates the controller's ability to prioritize and achieve high-efficiency kinematics without compromising thrust-following accuracy.

\RED{To further clarify the role of each cost component in (17), Table III reports an ablation study averaged over the three forward velocities (0.2, 0.3, 0.4  m/s), over a grid of target force commands. The results were averaged over all conditions tested after excluding the first control step to remove initialization effects. Table III summarizes the results using the following metrics: Force Error denotes the average force tracking error, St Deviation is its absolute deviation from the target value 0.30, In-Range \% indicates the percentage of cases falling within the acceptable Strouhal range [0.25, 0.35], Command Stability measures command variation between successive updates. When the Strouhal deviation increases from 0.0285 to 0.1227, the in-range ratio drops from 88.0\% to 34.7\%. When both the Strouhal and amplitude penalties are removed, the force tracking error becomes slightly smaller (0.1323 N), but this is achieved at the cost of substantially poorer efficiency, with only 30.0\% of the operating points remaining within the target Strouhal band. This indicates that force tracking alone does not lead to energetically favorable flapping patterns. In contrast, the amplitude penalty plays a complementary role by slightly improving the power consumption and command stability once the Strouhal term is present, while the smoothness term primarily suppresses abrupt command changes, reducing the command stability from 2.177 to 1.625. Overall, integration of all components provides the best balance between thrust tracking, Strouhal regulation, and executable smooth motion.}

A benchmark test was conducted to quantify the efficacy of the two-stage optimization algorithm, which combines a sampling-based MPC for broad exploration with a NMPC for local refinement. For this test, performance was assessed by comparing the final cost and computation time with and without the NMPC refinement stage. The benchmark covered 121 unique target force vectors, covering x components from 0.0 to -1.0 N and y components from 0.0 to 1.0 N in increments of 0.1 N. Each vector was tested for 5 cycles, totaling 605 trials, all at a fixed forward velocity of 0.3 m/s. As shown in Table \ref{tab:mpc_benchmark_updated}, the results indicate that the inclusion of the NMPC refinement stage consistently lowers the average cost of the solution, although at the expense of increased computation time.  Across the range of candidates tested, the NMPC stage reduced the final average cost by approximately 10-15\% compared to the sampling-based search alone. This result shows a clear trade-off, but multithreading allows us to achieve calculations at over 25Hz for each fin, which is still sufficient since the physical fin vibration frequency is around 3Hz in the maximum case.

\begin{figure*}[t]
\includegraphics[width=\linewidth]{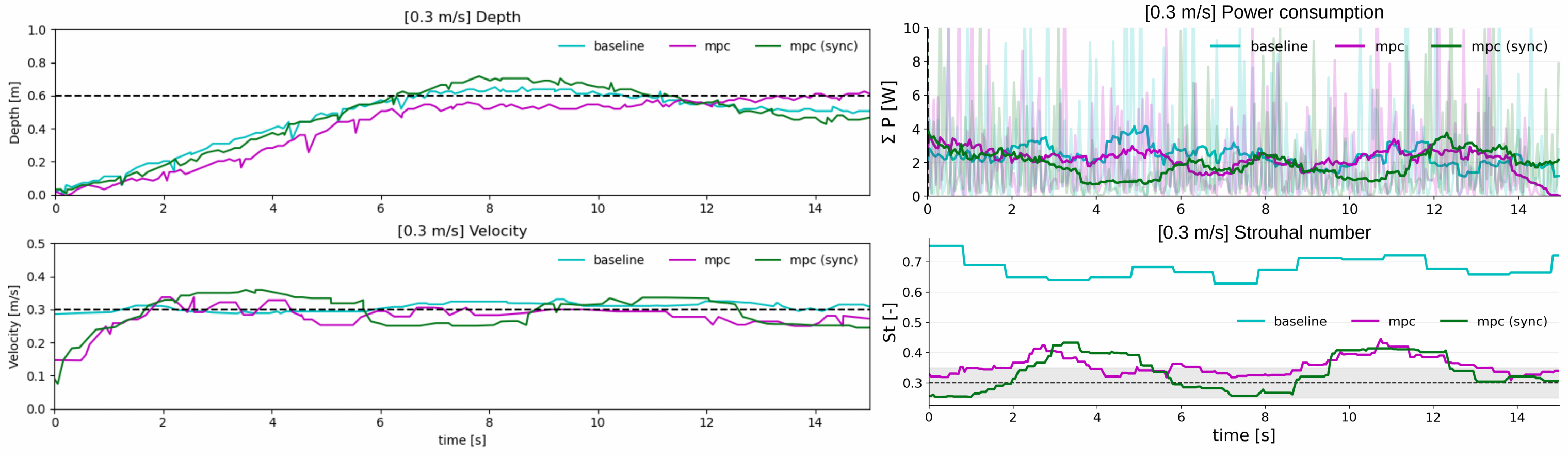}
\caption{\RED{Controller performance comparison at the target forward velocity of $0.3~m/s$. The panels display (from top to bottom, left to right) depth tracking, velocity tracking, total mechanical power consumption, and Strouhal number. While overall performance remains comparable across methods, the MPC controllers explicitly regulate the Strouhal number within the optimal range (shaded area), leading to a substantial reduction in total mechanical power compared to the baseline($\sum$P).}}
\label{energy}
\end{figure*}

\subsection{Pool experiment results}
A pool experiment was conducted to compare the efficiency performance of this controller. The experiments were carried out in a swimming pool with a depth of water of 1.65 m. The AUV was equipped with a pressure sensor for precise depth measurement and an inertial measurement unit (IMU) for attitude estimation. As illustrated in Fig. \ref{setup}.C, the experimental setup also included a DVL mounted at the rear of the vehicle to acquire velocity data. Before each experiment, the AUV was adjusted to achieve neutral buoyancy. The velocity signal was median-filtered and the pose was estimated by the IMU with an extended Kalman filter (EKF). For the trials, the proposed MPC controller (in both asynchronous and synchronized modes for the rear fins, explained in Section III.D) was compared with a baseline controller using an inverse model. The body frame velocity controller is exactly the same for all three setups described in Section III. The three tests were performed at each target body frame velocity to collect performance data (36 total), including vehicle state and mechanical power consumption. The instantaneous mechanical power consumed by each fin was estimated from the position data recorded on the encoder. This was achieved by first deriving the angular velocity ($\omega$) and the angular acceleration ($\alpha$) for each fin. The total power was then calculated as the sum of the inertial power ($P_{\text{inert}} = I \cdot \alpha \cdot \omega$), required to overcome the fin's own inertia, and the hydrodynamic drag power($P_{\text{drag}} = \tfrac{1}{2}\rho C_D A_{\text{f}} v^3$).
The experimental results demonstrate that the proposed MPC frameworks achieve energy savings. \RED{As a representative case, Fig. 7 shows the time-series results at a target forward velocity of 0.3 m/s. All controllers successfully tracked the target depth and forward velocity. Although the baseline achieved a slightly lower velocity Mean Absolute Error (MAE) at 0.3 m/s, the proposed MPC controllers maintained the Strouhal number closer to the optimal range and reduced the total mechanical power consumption. Table V shows the mean result of the experiments. The largest power reduction was observed at 0.2 m/s. This is consistent with the Strouhal statistics in Table V, where the inverse-model baseline shows its largest St error at 0.2 m/s, indicating that this operating point deviates the most from the efficient flapping corridor. This result corresponds to the large error observed in the simulation at 0.2 m/s in left panel of Fig. \ref{stanalysis}. Consequently, the proposed controller improved efficiency by regulating the fins toward the target Strouhal range.} 

\subsection{Field experiment results}
\begin{figure*}[t]
\includegraphics[clip, width=18cm]{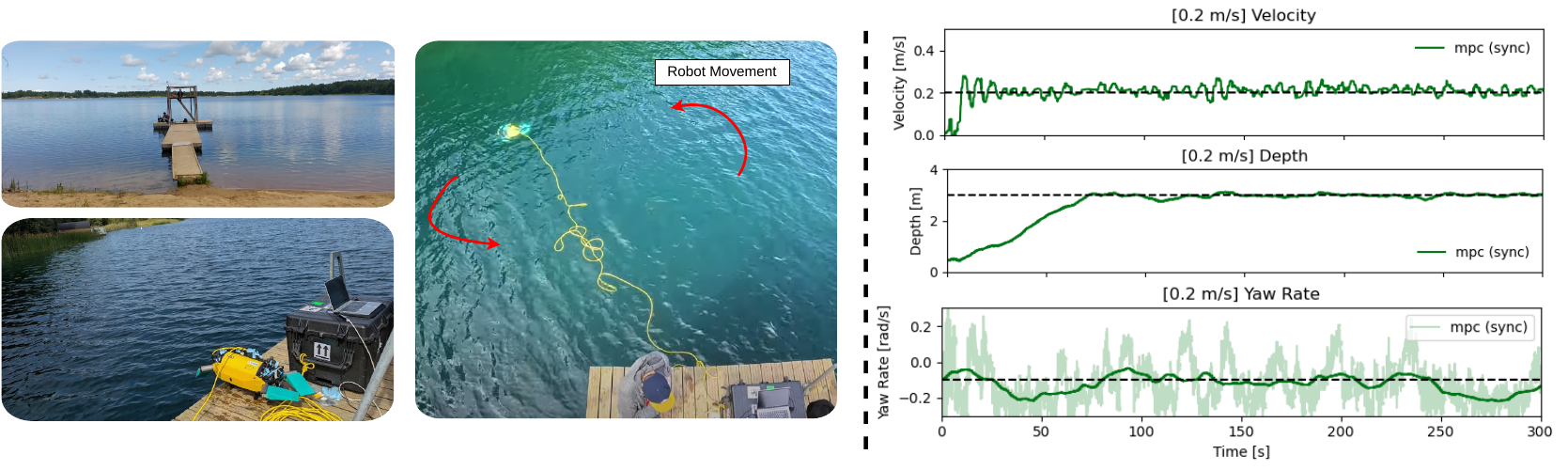}
\caption{ Field experiment and control performance validation in a real-world environment. The left panels show the experimental site at Rummu Lake. The right panels display 300 seconds of tracking data as the robot executed a circular trajectory with a target velocity of 0.2 m/s, a depth of 3.0 m, and a yaw rate of 0.1 rad/s. In each plot, the solid line represents the measured state, while the dashed line indicates the target setpoint.}
\label{field}
\end{figure*}

Experiments were conducted to verify the long-term performance of the proposed controller in a field environment. The experiments were carried out on Rummu Lake in Estonia. A controller with a synchronized rear fin was used, similar to the setup of the pool experiment. The wind velocity was 3.0 m/s average. The reference depth was set at 3.0 m, the target velocity was set at 0.2 m / s, and the target yaw rate was set at 0.1 rad / s, forming a large circle as shown on the left side of Fig. \ref{field}. These experiments were conducted for 300 seconds. The results are shown on the right side of Fig. \ref{field}. These results show that depth control increased linearly in the first half of the experiment and then successfully tracked the target value. The velocity also fluctuated with an error of about +/-0.025, which is similar to the results in the field environment as seen in the pool experiment. The yaw rate exhibited significant fluctuations, but the robot generally tracked the target value, moving in a large counterclockwise circle.

\section{Conclusion}

In this paper, we presented a novel control framework for multi-fin underwater robots that leverages a fundamental principle of efficient flapping propulsion to enhance endurance. We formulate a Model Predictive Control (MPC) scheme that explicitly incorporates a penalty for deviating from the optimal Strouhal number range of 0.25–0.35, thus steering the system toward motion patterns with high hydrodynamic efficiency. To solve the resulting nonconvex optimization problem under real-time constraints, we engineered a hybrid two-stage algorithm running onboard at 25 Hz. This method synergizes a broad sampling-based search for a starting approximated solution with a fast gradient-based search for local refinement. \RED{The performance of our approach was validated through comprehensive experiments. Ablation studies further confirmed that the Strouhal-aware MPC component contributes to maintaining oscillatory motion within the efficient range.} The pool experiments revealed significant reductions in mean mechanical power, ranging from 8.8\% to 32\% in a cruise range of 0.1 to 0.3 m/s relative to a conventional controller. Subsequent field experiments at Rummu Lake further underscored the controller's robustness and practical applicability, demonstrating stable trajectory tracking in a dynamic, real-world setting. \RED{These results indicate that regulating the Strouhal number within its optimal range improves the energy efficiency of multi-fin underwater robots.} Our future research will aim to integrate the optimal force allocation from the body-frame controller to develop an energy-sensitive control system.



%






\bibliographystyle{IEEEtran}
\bibliography{IEEEabrv,references}

\begin{IEEEbiography}[{\includegraphics[width=1in,height=1.2in,clip,keepaspectratio]{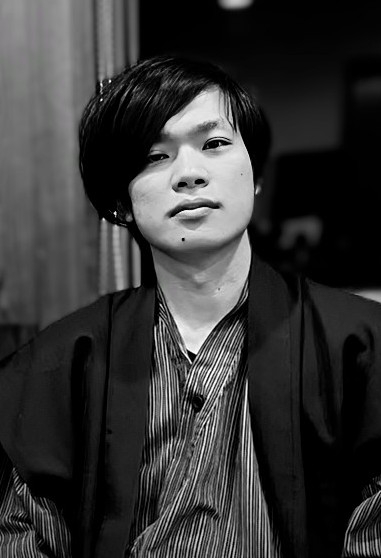}}]{Yuya Hamamatsu} received his M.Sc. degree in environment from the University of Tokyo, Japan, in 2020. He is currently working toward the Ph.D. degree with the Centre for Biorobotics at Tallinn University of Technology, Estonia. His research interests control theory on robotics. His research interests include robot policy learning, marine robotics, and robotic software architecture.
\end{IEEEbiography}

\begin{IEEEbiography}[{\includegraphics[width=1in,height=1.2in,clip,keepaspectratio]{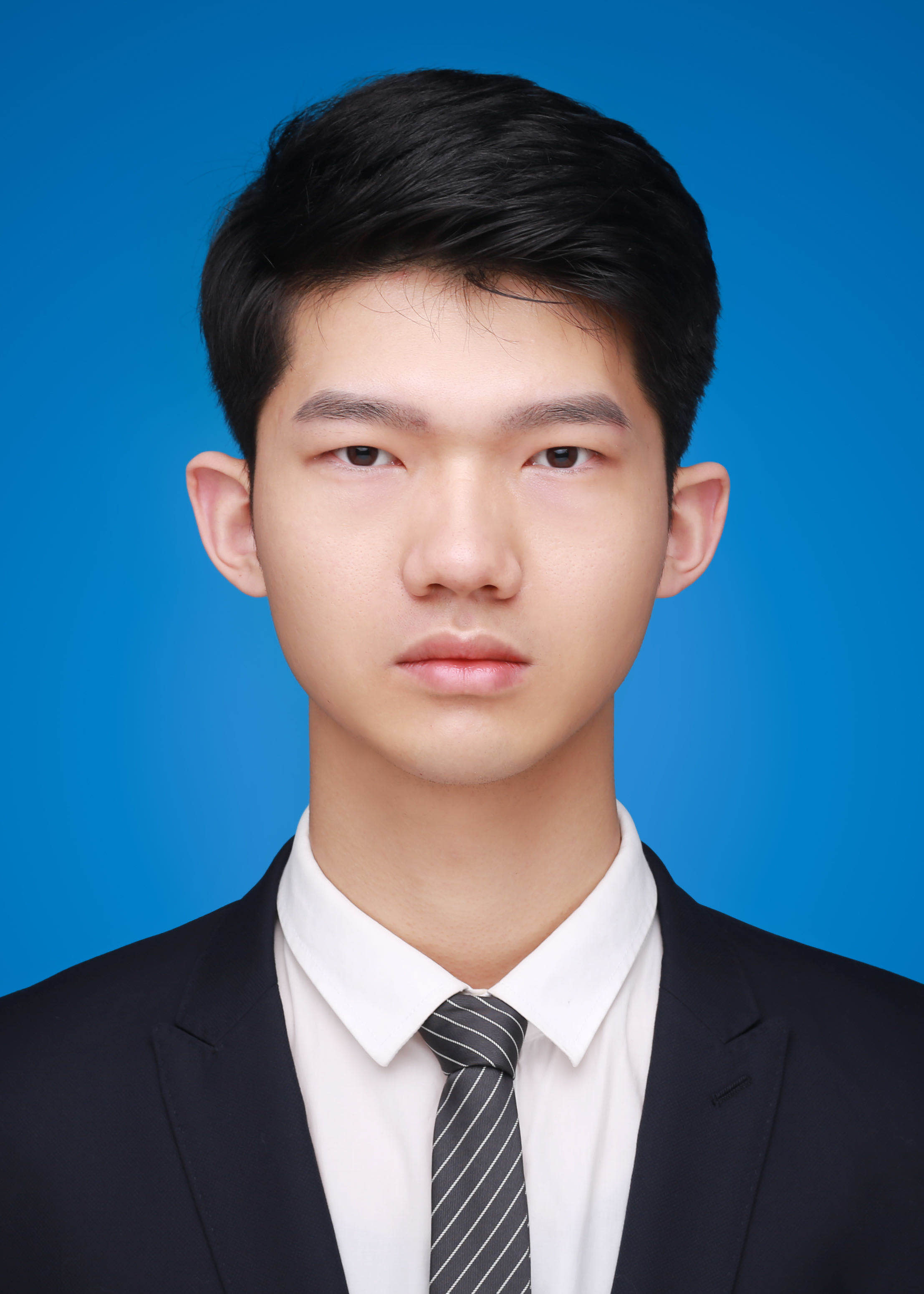}}]{Zixi Chen} received the M.Sc. degree in Control Systems from Imperial College in 2021 and the Ph.D. degree in Biorobotics from Scuola Superiore Sant’Anna. 

Dr. Chen is currently a Postdoctoral Researcher with the Brubotics, Vrije Universiteit Brussel and IMEC. 
His research interests include continuum robot control, surgical robotics, and optical tactile sensing.
\end{IEEEbiography}

\begin{IEEEbiography}[{\includegraphics[width=1in,height=1.2in,clip,keepaspectratio]{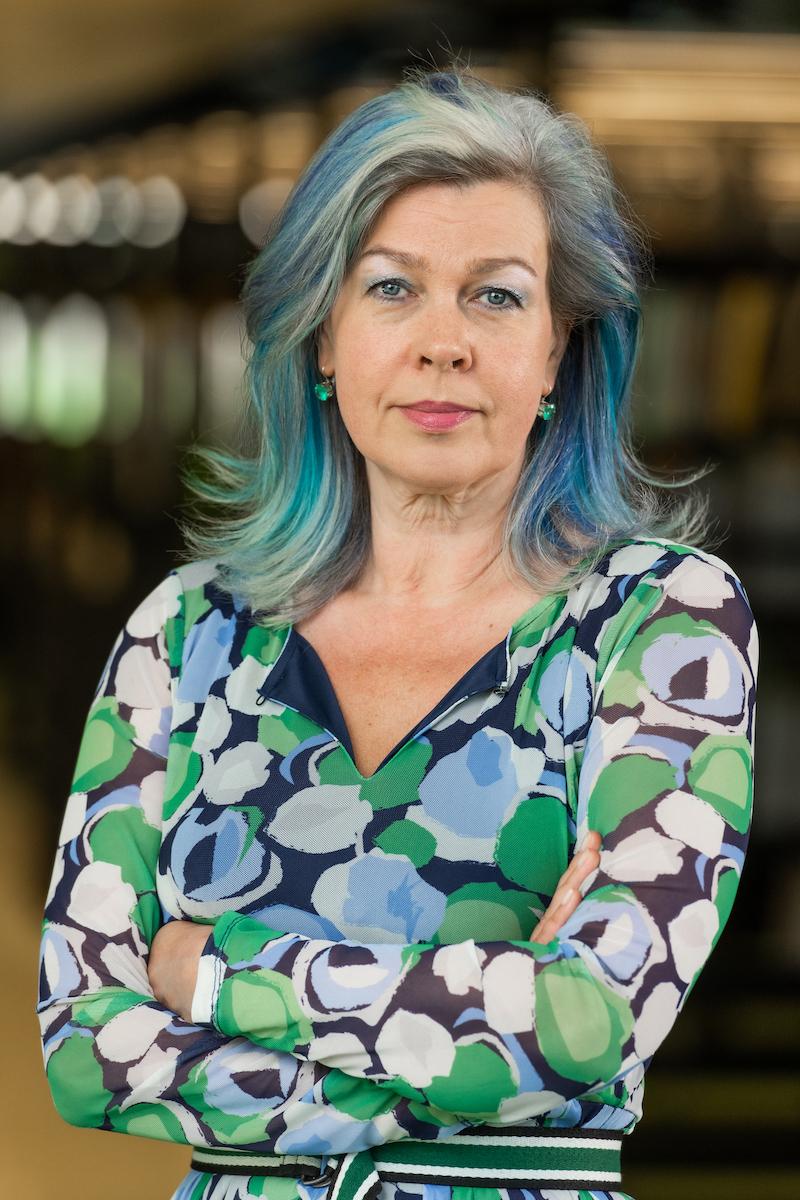}}]{Maarja Kruusmaa}
 received her Ph.D.from Chalmers University of Technology in 2002 and since 2008 is a professor in Tallinn University of Technology (TalTech) as a PI of Centre for Biororobitcs, a research group focusing on bio-inspired robotics, underwater robotics and novel underwater sensing technologies. 2017 - 2022 she was a visiting professor in NTNU AMOS (Centre for Excellence of Autonomous Marine Operations and Systems). Her research interests include novel locomotion mechanisms for underwater environments and on flowable media and novel methods for underwater flow sensing. 
 \end{IEEEbiography}

\begin{IEEEbiography}[{\includegraphics[width=1in,height=1.2in,clip,keepaspectratio]{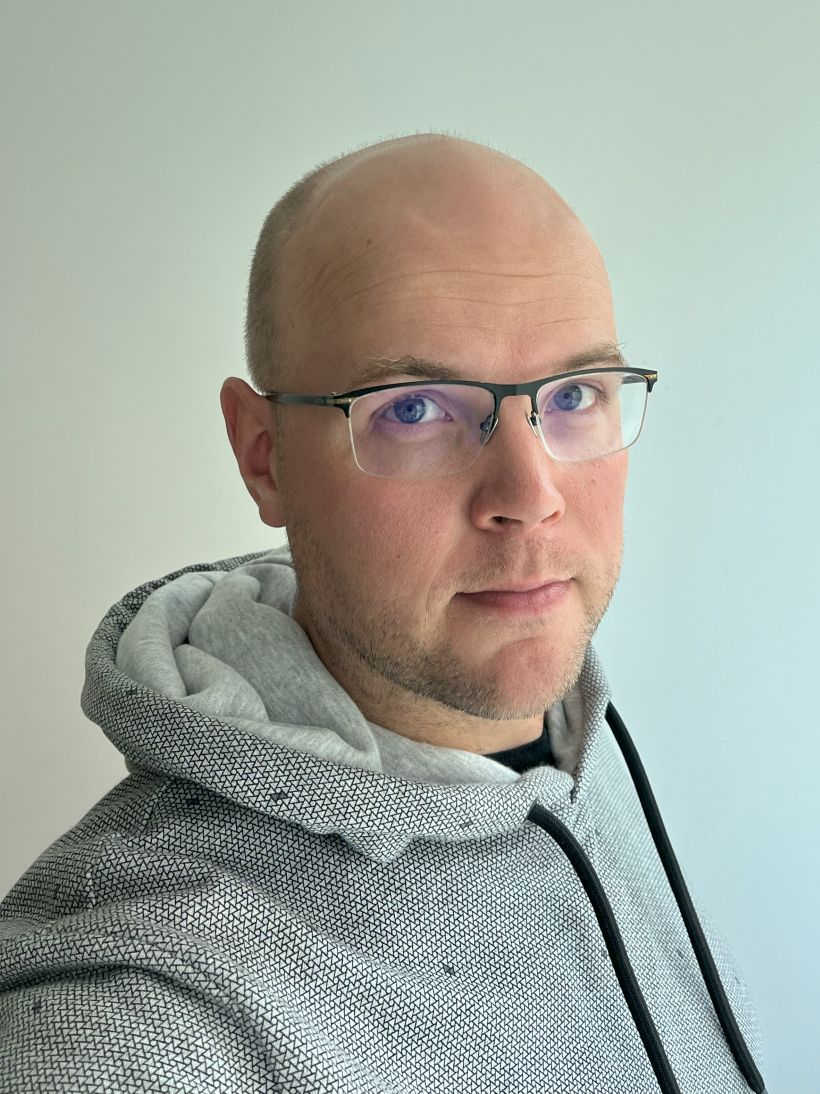}}]{Asko Ristolainen}
 received the B.Sc. and M.Sc. degrees in mechatronics and the Ph.D. degree in information and communication technology from the Tallinn University of Technology, Tallinn, Estonia, in 2008, 2010, and 2015, respectively.
He is currently a Senior Researcher with the Centre for Biorobotics, School of Information Technologies, Tallinn University of Technology.
His research interests include tactile and flow sensor design in robotics and remote environmental sensing.
\end{IEEEbiography}

\end{document}